\documentclass[10pt,twocolumn,letterpaper]{article}

\usepackage[pagenumbers]{cvpr} 

\usepackage[dvipsnames]{xcolor}

\definecolor{cvprblue}{rgb}{0.21,0.49,0.74}
\usepackage[pagebackref,breaklinks,colorlinks,citecolor=cvprblue]{hyperref}

\usepackage{bm}
\usepackage{multirow}
\usepackage{adjustbox}
\usepackage{pifont}
\usepackage{color, colortbl}
\definecolor{Gray}{gray}{0.9}
\usepackage[capitalize]{cleveref}
\crefname{section}{Sect.}{Secs.}
\crefname{table}{Tab.}{Tabs.}
\usepackage{subcaption}
\usepackage{nccmath}

\newcommand{\cmark}{\ding{51}}%
\newcommand{\xmark}{\ding{55}}%
\newcommand{\ieno}{\textit{i}.\textit{e}.}
\newcommand{\egno}{\textit{e}.\textit{g}.} 

\def\confName{CVPR}
\def\confYear{2024}

\title{ReGenNet: Towards Human Action-Reaction Synthesis}

\newcommand*{\affmark}[1][*]{\textsuperscript{#1}}

\author{
Liang Xu\affmark[1,2] \quad
Yizhou Zhou\affmark[3] \quad
Yichao Yan\affmark[1$\dagger$] \quad
Xin Jin\affmark[2$\dagger$] \quad
Wenhan Zhu\affmark[1] \quad
Fengyun Rao\affmark[3] \\
Xiaokang Yang\affmark[1] \quad
Wenjun Zeng\affmark[2] \quad
\vspace{0.7em} \\
\affmark[1]{MoE Key Lab of Artificial Intelligence, AI Institute, Shanghai Jiao Tong University, Shanghai, China} \\
\affmark[2]{Ningbo Institute of Digital Twin, Eastern Institute of Technology, Ningbo, China} \\
\affmark[3]{WeChat, Tencent Inc.} \\
{\small\url{https://liangxuy.github.io/ReGenNet/}}
\vspace{-0.5em}
}

\begin{document}
\maketitle

\let\thefootnote\relax\footnotetext{$^\dagger$Corresponding authors}

\begin{abstract}
Humans constantly interact with their surrounding environments.
Current human-centric generative models mainly focus on synthesizing humans plausibly interacting with \textit{static} scenes and objects, while the \textit{dynamic} human action-reaction synthesis for ubiquitous causal human-human interactions is less explored. Human-human interactions can be regarded as asymmetric with actors and reactors in atomic interaction periods.
In this paper, we comprehensively analyze the asymmetric, dynamic, synchronous, and detailed nature of human-human interactions and propose the \textit{first} multi-setting human action-reaction synthesis benchmark to generate human reactions conditioned on given human actions.
To begin with, we propose to annotate the actor-reactor order of the interaction sequences for the NTU120, InterHuman, and Chi3D datasets.
Based on them, a diffusion-based generative model with a Transformer decoder architecture called \textbf{ReGenNet} together with an explicit distance-based interaction loss is proposed to predict human reactions in an online manner, where the future states of actors are unavailable to reactors.
Quantitative and qualitative results show that our method can generate instant and plausible human reactions compared to the baselines, and can generalize to unseen actor motions and viewpoint changes.
\end{abstract}

\vspace{-3mm}
\section{Introduction}
\begin{figure*}[t]
  \vspace{-2mm}
  \centering
  \includegraphics[width=1.0\linewidth]{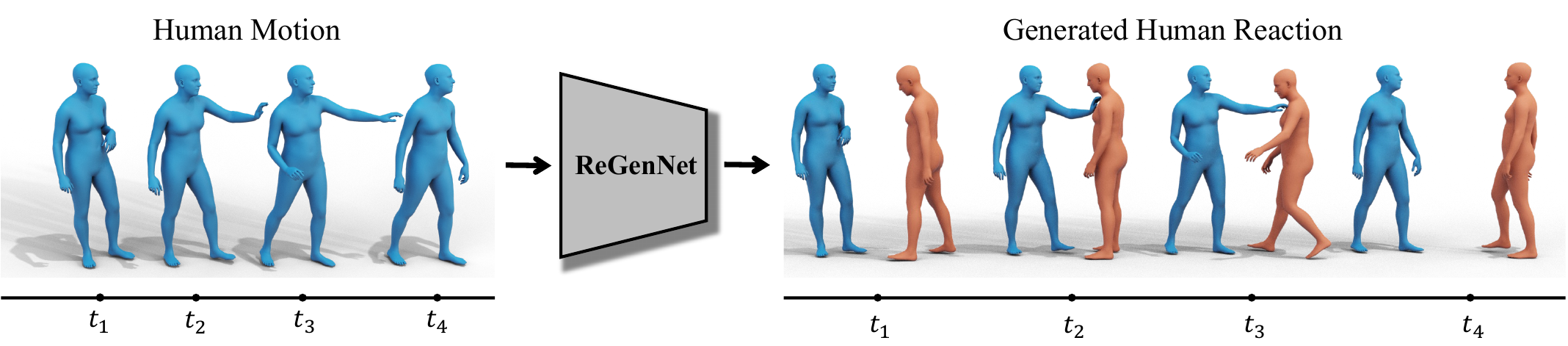}
  \vspace{-2mm}
  \caption{Illustration of our proposed \textbf{ReGenNet}, \ieno, given a human motion sequence and generate the plausible human reactions, which will have broad applications in AR/VR and games.}
  \label{fig:intro}
  \vspace{-5mm}
\end{figure*}

\vspace{-2mm}
Human-centric generative models have been widely studied with numerous applications. 
Currently, there exists substantial progress on generative models to synthesize how digital humans \textit{actively} interact with the environments with physical and semantic plausibility, \egno, conditioned on a given scene~\cite{zhang2020generating,place,hassan2021populating,wang2021synthesizing,hassan2021stochastic,zhao2022compositional} and object~\cite{grab,saga,toch,goal}.
However, for human-human interactions, a man could be \textit{active} or \textit{passive} in atomic interaction periods. Existing works for human motion generation mainly treat the actors and reactors equally or limited on single human motion generation~\cite{actformer,shafir2023human,intergen}, while neglecting the reaction generation problem for ubiquitous human-human interactions (see ~\cref{fig:intro}).
In this paper, we focus on generative models for human action-reaction synthesis, \ieno, generating human reactions given the action sequence of another as conditions.
We believe this task will contribute to many applications in AR/VR,  games, human-robot interaction, and embodied AI.

Modeling human-human interactions is a challenging task with the following features: 
1) \textbf{Asymmetric}, \ieno, the actor and reactor play asymmetric roles during a causal interaction, where one person acts, and the other reacts~\cite{yun2012two};
2) \textbf{Dynamic}, \ieno, during the interaction period, the two people constantly wave their body parts, move close/away, and change relative orientations, spatially and temporally; 
3) \textbf{Synchronous}, \ieno, typically, one person responds instantly with others such as an immediate evasion when someone throws a punch, thus the online generation is required;
4) \textbf{Detailed}, \ieno, the interaction between humans involves not only coarse body movements together with relative position changes but also local hand gestures and even facial expressions. 
Thus, it is desirable to design a generative model that simultaneously considers the above characteristics.

Directly applying previous human-centric generative models~\cite{zhang2020generating,place,toch,goal} for human action-reaction synthesis is impractical, because existing models typically consider \textit{static} scenes or objects, yet \textit{dynamic} humans are more complicated. 
Moreover, online generation is also not required for human scene/object interaction scenarios, yet significant for action-reaction synthesis.
On the other hand, recent years have witnessed the rapid development of single human motion generation conditioned on action categories~\cite{actor,mdm}, text descriptions~\cite{temos,guo2022generating,flame}, audios~\cite{dance2music,li2021ai,li2022danceformer,habibie2022motion,ao2022rhythmic} or sparse tracking signals~\cite{jiang2022avatarposer,agrol,aliakbarian2022flag}.
However, very few works~\cite{actformer,intergen,starke2021neural,starke2020local,xu2023inter} have been proposed to generate multi-person interactions, yet treating the actor-reactor equally~\cite{actformer,intergen} or focus on specific action categories, such as ``martial arts''~\cite{starke2021neural}.
The sparse skeleton joints or SMPL body models~\cite{smpl} are widely adopted, while greatly limiting the fineness and \textbf{details} of hands-involved interactions such as ``playing finger guessing'', letting alone the \textbf{synchronous} and \textbf{asymmetric} natures.
To the best of our knowledge, no previous works have been proposed to deal with all the aforementioned patterns of human-human interactions.
There are no such human-human interaction datasets with actor-reactor annotations.

In this paper, we comprehensively consider the intrinsic features of human-human interactions and propose the first human action-reaction synthesis benchmark with the following designs:
1) We adopt the SMPL-X~\cite{smplx} body model as our data representation because it contains \textbf{detailed} articulated hand poses. In terms of the datasets, we choose Chi3D~\cite{chi3d} with SMPL-X annotations from the body markers; we also extend the widely used NTU120 dataset~\cite{nturgbd120} to SMPL-X version by a state-of-the-art pose estimation method~\cite{pymaf-x}; we also adopt the human-human interaction MoCap dataset InterHuman~\cite{intergen} for its accurate motion sequences. 
2) For the absence of \textbf{asymmetry} nature in current interaction datasets, we annotate the actors and reactors of the above three datasets.
Based on these annotations, we propose, to our best knowledge, the \textit{first} multi-setting human action-reaction synthesis benchmark aiming to generate physically and semantically plausible human reactions conditioned on a given person's action sequence.
3) To generate instant and \textbf{synchronous} human reactions, we need to design an \textit{online} model, \ieno, future human motion is unavailable for the synthesis at the current moment. We adopt a diffusion model together with the Transformer architecture to model the spatiotemporal interactions, and we choose the Transformer-decoder for its leftward property via the masked multi-head attention and inference in an auto-regressive manner.
4) To handle the highly \textbf{dynamic} human-human interactions, we draw inspiration from the previous human scene/object interaction counterparts which model the contact/interaction using distance-based representations~\cite{place,goal}.
We thus design interaction losses that explicitly measure the relative distances of the interacted spatiotemporal body poses, orientations, and translations. 
Considering that in practical applications, the intention of the actor could be agnostic to reactors, we also train our model in an unconstrained fashion~\cite{modi,mdm}.
With the above designs, we name our reaction generation model as \textbf{ReGenNet}. 
Extensive experiments show that ReGenNet can synthesize realistic human reactions with the lowest time delay compared to the baselines, and can generalize to unseen actor motions and viewpoint changes. Our model is modular and flexible to be trimmed for other practical applications such as multi-person interaction generation tasks.

Our contributions can be summarized as follows.
We comprehensively analyze the asymmetric, dynamic, synchronous, and detailed nature of human-human interactions. 
Based on these analyses, we propose the first multi-setting human action-reaction synthesis benchmark with three dedicated annotated datasets. 
To address this task, we present \textbf{ReGenNet}, a diffusion-based generative model to synthesize plausible and instant human reactions.
Our benchmark, data, models, and code will be made publicly available.

\section{Related Work}
\vspace{-2mm}
\noindent{\textbf{Human-scene/object interaction.}}
Synthesizing human-scene/object interactions is critical for games and AR/VR applications~\cite{yan_survey}.
The goal is to fit the human body with scenes/objects as \textit{contexts}, so as to plausibly navigate/interact in the scenes or manipulate the objects with geometric and semantic constraints.
For \textit{human-scene interactions},~\cite{grabner2011makes,gupta20113d,kim2014shape2pose,savva2014scenegrok,savva2016pigraphs,li2019putting,hu2020predictive,zhang2020generating,place,hassan2021populating} can generate static interactions with unseen environments. Recent works~\cite{wang2021synthesizing,hassan2021stochastic,zhao2022compositional} extended to produce dynamic human-scene interactions, which is equivalent to generating motion sequences with scene contexts.
For \textit{human-object interactions}, earlier works~\cite{karunratanakul2020grasping,grady2021contactopt} focused on generating hand-object interactions.
\cite{hsiao2006imitation,borras2015whole,grab} built databases of whole-body interactions with daily objects and promoted the research works on synthesizing whole-body manipulations with objects~\cite{grab,saga,toch,goal}.

The core of these human-centric generative models lies in understanding the semantics and affordances of scenes/objects, together with training a conditional generative model based on the scene/object priors.
Our proposed benchmark of human action-reaction synthesis, where the motion of the actor can also be viewed as the \textit{contexts}, is also significant yet not explored to the best of our knowledge. The highly dynamic human motion is also more complicated than static scenes and objects.

\noindent{\textbf{Human motion generation.}}
The goal is to generate human motion conditioned on different guidances.
Early works~\cite{fragkiadaki2015recurrent,martinez2017human,hernandez2019human,mao2022weakly,guo2023back,barquero2022belfusion,tang2023predicting} focused on the task of future motion prediction, \ieno, to predict the motion of future frames given past motion as guidance.
Besides, motion synthesis from high-level semantic signals such as action labels~\cite{action2motion,csgn,actformer,actor,mdm,cervantes2022implicit,chen2022executing}, text~\cite{language2pose,motiondiffuse,temos,guo2022generating,flame,mofusion}, music~\cite{dance2music,li2021ai,aristidou2021rhythm,li2022danceformer}, speech~\cite{habibie2022motion,ao2022rhythmic} have emerged in recent years. 
However,~\cite{modi,mdm} also proved that human motion generation can also be learned without conditions in an unsupervised manner.
Furthermore, human-human interaction synthesis has also been noticed~\cite{actformer,shafir2023human,intergen,starke2021neural,starke2020local}. However, these works treated the actor-reactor equally~\cite{actformer,intergen} or focused on specific action categories~\cite{starke2021neural} in graphics.
Another line of research in AR/VR scenarios reverted the full-body poses from the sparse tracking signal of head and wrists~\cite{huang2018deep,jiang2022avatarposer,agrol,aliakbarian2022flag,castillo2023bodiffusion}, facilitating real-world applications.

The key component of these works is to learn generative models such as GANs~\cite{gan,csgn,actformer}, VAEs~\cite{vae,action2motion,actor,cervantes2022implicit}, flow-based models~\cite{rezende2015variational,aliakbarian2022flag} and diffusion models~\cite{ddpm,motiondiffuse,mdm,mofusion}. In this paper, we also adopt the diffusion models for high-quality synthesis.
However, most of previous works neglected the asymmetry of causal human-human interactions. 
Concurrent works~\cite{chopin2023interaction,role_aware} target at human reaction generation, yet~\cite{chopin2023interaction} adopts very sparse skeleton representations and~\cite{role_aware} only handles the ``offline'' and ``unconstrained'' setting of human reaction generation without generating instant and intention agnostic reactions.
\begin{table}[t]
\begin{center}
\resizebox{\linewidth}{!}{
\begin{tabular}{ l c c c c c}
\toprule
Dataset & Year & Verbs & Motions & Modality & Asymmetry\\
\midrule
SBU~\cite{yun2012two} & 2012 & 8 & 300 & Skel. & \xmark \\
ShakeFive2~\cite{van2016spatio} & 2016 & 8 & 153 & Skel. & \xmark \\
K3HI~\cite{baruah2020multimodal} & 2020 & 8 & 312 & Skel. & \xmark \\
NTU120~\cite{nturgbd120} & 2019 & 26 & 8,276 & Skel. & \xmark \\
You2Me~\cite{ng2020you2me} & 2020 & 4 & 42 & Skel. & \xmark\\
Chi3D~\cite{chi3d} & 2020 & 8 & 373 & SMPL-X & \xmark\\
InterHuman~\cite{intergen} & 2023 & - & 6,022 & SMPL & \xmark \\
\midrule
Chi3D-AS(Ours) & 2023 & 8 & 373 & SMPL-X & \cmark \\
InterHuman-AS(Ours) & 2023 & - & 6,022 & SMPL & \cmark \\
NTU120-AS(Ours) & 2023 & 26 & 8,118 & SMPL-X & \cmark \\
\bottomrule
\end{tabular}
}
\end{center}
\vspace{-4mm}
\caption[caption]{\textbf{Human-human interaction datasets}. Skel. denotes skeleton and AS denotes asymmetry.}
\label{tab:dataset}
\end{table}

\noindent{\textbf{Human-human interaction dataset.}}
Human-human interactions are indispensable components in our daily lives. Many multi-person interaction datasets such as UMPM~\cite{van2011umpm}, SBU~\cite{yun2012two}, ShakeFive2~\cite{van2016spatio}, K3HI~\cite{baruah2020multimodal}, You2Me~\cite{ng2020you2me}, Chi3d~\cite{chi3d}, NTU120~\cite{nturgbd120}, ExPI~\cite{expi}, InterHuman~\cite{intergen} have been produced with various sizes and modalities as in~\cref{tab:dataset}. Especially, NTU120~\cite{nturgbd120} is a large-scale human motion dataset with 26 interactive action categories and concurrently, InterHuman~\cite{intergen} brings the currently largest multi-human interaction dataset with text description annotations.

However, all these previous datasets overlooked the asymmetry property of causal human-human interactions. Thus, we propose to annotate the actor-reactor order of the Chi3D, NTU120 and InterHuman datasets. We also extend the NTU120 dataset to the SMPL-X version to better describe the fine-grained interaction patterns.

\section{Human Action-Reaction Synthesis}
\vspace{-2mm}
\subsection{Modeling setup}
\vspace{-2mm}
\textbf{Dataset Formulation.}
In this work, we tackle the problem of fine-grained human action-reaction synthesis. However, we notice that all the previous multi-person interaction datasets ignored the asymmetry property of causal relationships (see~\cref{tab:dataset}). Thus, we choose three datasets, \ieno, NTU120~\cite{nturgbd120}, InterHuman~\cite{intergen} and Chi3D~\cite{chi3d}, and annotate the actor-reactor order of each interaction sequence. We also extend the NTU120 dataset to SMPL-X representation by a pose estimation method~\cite{pymaf-x}. We name the datasets as NTU120-AS, InterHuman-AS and Chi3D-AS, where ``AS'' denotes asymmetry. For the details of the asymmetric action definition and the labeling process, please refer to the supplementary materials.
\begin{figure*}[t]
  \vspace{-2mm}
  \centering
  \includegraphics[width=1.0\linewidth]{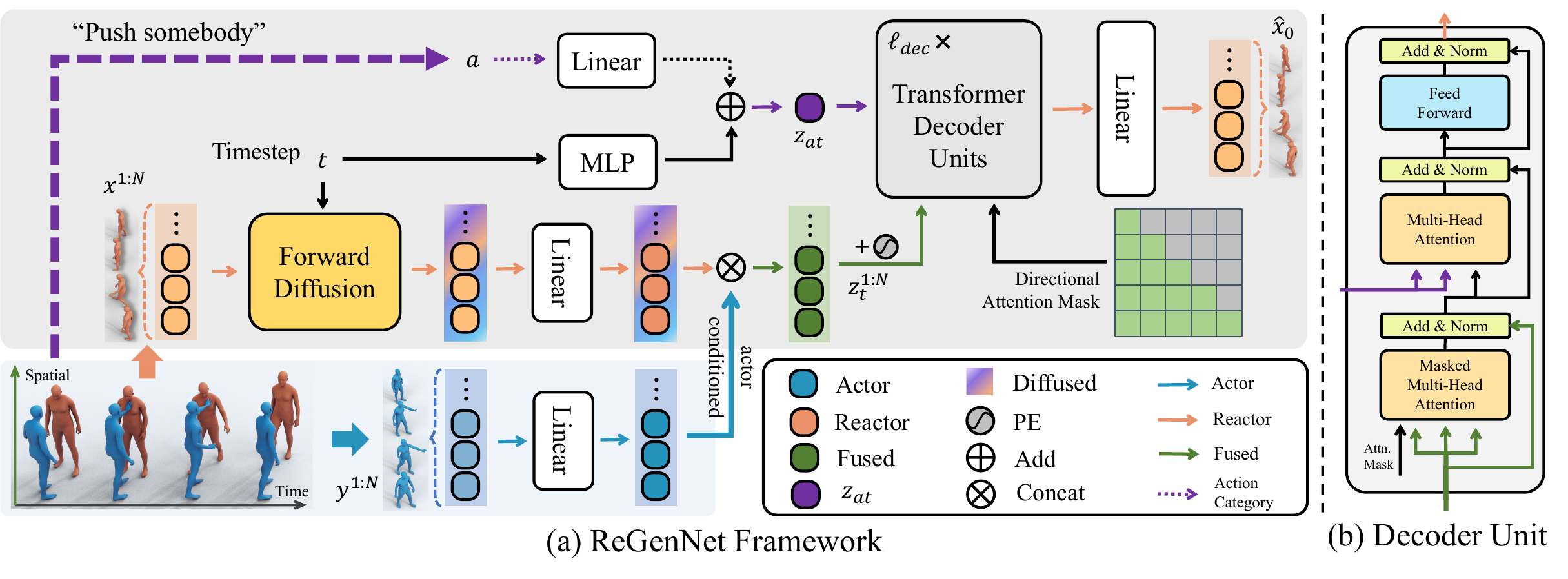}
  \vspace{-2mm}
  \caption{\textbf{The architecture} of our proposed \textbf{ReGenNet} which is formulated in a diffusion-based framework with Transformer Decoder Units. The \textcolor{gray}{gray} panel of (a) illustrates the whole diffusion model with the ``Forward Diffusion'' process and a stack of $\ell_{dec}$ ``Transformer Decoder Units'' as the denoising process, the \textcolor{blue}{blue} panel of (a) is the actor feature as the condition. (b) shows the details of the ``Transformer Decoder Units'' with directional attention mask for \textit{online} reaction synthesis.}
  \label{fig:pipeline}
  \vspace{-2mm}
\end{figure*}

\noindent\textbf{Problem Formulation.}
In the setting of human action-reaction synthesis, our goal is to generate the reaction conditioned on an arbitrary action.
Formally, we denote the reaction as $x^{1:N}=\{x^i\}_{i=1}^N$ and the action as $y^{1:N} = \{y^i\}_{i=1}^N$ of duration $N$. The intention $a$ can be a signal of action label, text, and audio to dictate the interaction, which could be optional for intention-agnostic scenarios.

To enhance the representational power of human-human interactions, we adopt SMPL-X~\cite{smplx} human model to represent the human motion sequence.
Thus, the reaction can be represented as $x^i = \left[\bm{\theta}^x_i, \bm{q}^x_i, \bm{\gamma}^x_i\right]$
where $\bm{\theta}^x_i\in\mathbb{R}^{3K}$, $\bm{q}^x_i\in \mathbb{R}^3$, $\bm{\gamma}^x_i\in \mathbb{R}^3$ are the pose parameters, global orientation and the root translation of the person, respectively. $K=54$ is the number of body joints together with the jaw, eyeballs, and fingers.

\noindent\textbf{Motion Diffusion Model.}
Diffusion models~\cite{sohl2015deep,ddpm} have been proven to serve as a powerful generative model for human motion synthesis~\cite{motiondiffuse,mdm,mofusion,agrol}, which can be regarded as learning a Markov chain-based progressive noising and denoising of human motions. Given the reaction $x^{1:N}_0$ sampled from the real interaction data distribution, the noising process can be written as
\begin{equation}\label{eqn:diffusion}
{
q(x^{1:N}_{t} | x^{1:N}_{t-1}) = \mathcal{N}(x^{1:N}_{t}; \sqrt{\alpha_{t}}x^{1:N}_{t-1},(1-\alpha_{t})I),
}
\end{equation}
where $t$ is the timestep of the diffusion process, $\alpha_t \in (0, 1)$ is a constant hyper-parameter and $I$ is the identity matrix. With sufficiently large $T$, $\alpha_t$ becomes small enough and $x^{1:N}_T$ can be approximated as a Gaussian noise $\mathcal{N}(0, I)$. To generate the high-fidelity reaction, we need to reverse denoise the $x_T$ back to $x_0$ for $T$ timesteps.
In our setting, the reverse diffusion process is conditionally formulated as $p(x^{1:N}_{t-1}|x^{1:N}_t, y^{1:N}, a)$. We follow~\cite{ramesh2022hierarchical,mdm,agrol} to use a neural network $F$ to directly predict the clean body poses instead of predicting the residual noise, \ieno, $\hat{x}_0=F(x_t, y^{1:N}, t, a)$, so that we can add geometric losses directly on the predicted $\hat{x}_0$. $F$ can be implemented by Transformers or MLP networks for different applications.
The training objective of the diffusion model is formulated as
\begin{equation}\label{eqn:loss_diff}
{
\mathcal{L}_{dm} = \mathbb{E}_{x_0 \sim q(x_0), t \sim [1,T]}[\| x_0 - F(x_t, y^{1:N}, t, a)\|_2^2].
}
\end{equation}

\subsection{Reaction Generation Network}
\label{sec:regennet}
In this section, we introduce our holistic diffusion-based reaction generation framework as shown in~\cref{fig:pipeline}, which consists of a diffusion model $M$ and a Transformer decoder $F$. Given a coupled action-reaction pair and the optional action category (dotted lines in~\cref{fig:pipeline}) $<$$x^{1:N}$, $y^{1:N}$, $a$$>$, $y^{1:N}$ and $a$ serve as conditions and $x^{1:N}$ is the reaction to generate. For a sampled noising timestep $t$, we add noise to $x^{1:N}$ through the forward diffusion process as~\cref{eqn:diffusion} to produce the noised $x^{1:N}_t$. Then both the $x^{1:N}_t$ and the $y^{1:N}$ are linearly projected to obtain the latent features through an FC layer to dimension $d$, and concatenated together through the feature dimension. We also apply another FC layer to fuse the concatenated features and reduce the dimension to produce the final tokens $z^{1:N}_t$. Experimental results show that the conditioning scheme by concatenation is simple yet effective.
The noising timestep $t$ and the optional action label $a$ are all projected to dimension $d$ by separate feed-forward networks and then summed up to obtain the token $z_{at}$.

We implement $F$ by stacking $\ell_{dec}$ layers of Transformer decoder units to prevent future information leakage via the masked multi-head attention for \textit{online} generation. $F$ takes $z_{at}$ as input tokens and $z^{1:N}_t$ summed with a standard positional embedding as output tokens together with a directional attention mask, which is critical to prevent the model from seeing future actions at the current moment. The output of the decoders is then projected back as the predicted clean body poses $\hat{x}^{1:N}_0$. During inference time, we generate human reactions in an auto-regressive manner.

\begin{table*}[t]
  \begin{center}
  \resizebox{1.0\textwidth}{!}{
  \begin{tabular}{l c c c c c c c c }
  \toprule
  \multirow{2}{*}{Method} & \multicolumn{4}{c}{Train conditioned} & \multicolumn{4}{c}{Test conditioned} \\
  \cmidrule(rl){2-5} \cmidrule(rl){6-9}
  & FID$\downarrow$ & Acc.$\uparrow$ & Div.$\rightarrow$ & Multimod.$\rightarrow$ & FID$\downarrow$ & Acc.$\uparrow$ & Div.$\rightarrow$ & Multimod.$\rightarrow$ \\
  \midrule
  Real & $0.09^{\pm0.00}$ & $1.000^{\pm0.0000}$ & $10.54^{\pm0.06}$ & $26.71^{\pm0.62}$ & $0.09^{\pm0.00}$ & $0.867^{\pm0.0002}$ & $13.06^{\pm0.09}$ & $25.03^{\pm0.23}$ \\
  \midrule
  cVAE~\cite{vae} & $77.52^{\pm7.25}$ & $0.899^{\pm0.0002}$ & $10.10^{\pm0.02}$ & $19.38^{\pm0.16}$ & $70.10^{\pm3.42}$ & $\underline{0.724^{\pm0.0002}}$ & $11.14^{\pm0.04}$ & $18.4^{\pm0.26}$\\
  AGRoL~\cite{agrol} & $38.04^{\pm1.45}$ & $0.932^{\pm0.0001}$ & $10.95^{\pm0.07}$ & $21.44^{\pm0.34}$ & $44.94^{\pm2.46}$ & $0.680^{\pm0.0001}$ & $\underline{12.51^{\pm0.09}}$ & $19.73^{\pm0.17}$ \\
  MDM~\cite{mdm} & $40.13^{\pm3.65}$ & $0.955^{\pm0.0001}$ & $\mathbf{10.53^{\pm0.04}}$ & $21.15^{\pm0.26}$ & $54.54^{\pm3.94}$ & $0.704^{\pm0.0003}$ & $11.98^{\pm0.07}$ & $19.45^{\pm0.20}$ \\
  MDM-GRU~\cite{mdm} & $\underline{5.31^{\pm0.18}}$ & $\underline{0.993^{\pm0.0000}}$ & $11.03^{\pm0.06}$ & $\underline{25.04^{\pm0.36}}$ & $\underline{24.25^{\pm1.39}}$ & $0.720^{\pm0.0002}$ & $\mathbf{13.43^{\pm0.09}}$ & $\underline{22.24^{\pm0.29}}$\\
  \cellcolor{Gray}ReGenNet & \cellcolor{Gray}{$\mathbf{0.90^{\pm0.01}}$} & \cellcolor{Gray}{$\mathbf{1.000^{\pm0.0000}}$} & \cellcolor{Gray}$\underline{10.69^{\pm0.05}}$ & \cellcolor{Gray}$\mathbf{26.25^{\pm0.35}}$ & \cellcolor{Gray}$\mathbf{11.00^{\pm0.74}}$ & \cellcolor{Gray}$\mathbf{0.749^{\pm0.0002}}$ & \cellcolor{Gray}$13.80^{\pm0.16}$ & \cellcolor{Gray}$\mathbf{22.90^{\pm0.14}}$ \\
  \bottomrule
  \end{tabular}
  }
  \end{center}
  \vspace{-4mm}
  \caption[caption]{\textbf{Comparison to state-of-the-arts} on the \textit{online, unconstrained} setting for human action-reaction synthesis on NTU120-AS. $\pm$ indicates 95\% confidence interval, $\rightarrow$ means that closer to Real is better. \textbf{Bold} indicates best result and \underline{underline} indicates second best.}
  \label{tab:cmdm}
\end{table*}

\begin{table*}[t]
\vspace{-1mm}
\begin{center}
\resizebox{1.0\textwidth}{!}{
\begin{tabular}{l c c c c c c c c}
\toprule
\multirow{2}{*}{Method} & \multicolumn{4}{c}{Train conditioned} & \multicolumn{4}{c}{Test conditioned}\\
\cmidrule(rl){2-5} \cmidrule(rl){6-9}
& FID$\downarrow$ & Acc.$\uparrow$ & Div.$\rightarrow$ & Multimod.$\rightarrow$ & FID$\downarrow$ & Acc.$\uparrow$ & Div.$\rightarrow$ & Multimod.$\rightarrow$ \\
\midrule
Real & $0.19^{\pm0.01}$ & $1.000^{\pm0.0000}$ & $5.36^{\pm0.08}$ & $20.06^{\pm0.78}$ & $0.75^{\pm0.18}$ & $0.691^{\pm0.0093}$ & $7.15^{\pm1.27}$ & $12.94^{\pm0.96}$ \\
\midrule
cVAE~\cite{vae} & $25.45^{\pm13.9}$ & $0.843^{\pm0.0005}$ & $9.02^{\pm0.30}$ & $13.82^{\pm0.64}$ & $17.33^{\pm17.14}$ & $0.552^{\pm0.0024}$ & $8.20^{\pm0.57}$ & $\underline{11.44^{\pm0.35}}$ \\
AGRoL~\cite{agrol} & $47.73^{\pm5.95}$ & $0.975^{\pm0.0001}$ & $7.43^{\pm0.21}$ & $15.59^{\pm0.49}$ & $64.83^{\pm277.8}$ & $\underline{0.644^{\pm0.0039}}$ & $\mathbf{7.00^{\pm0.95}}$ & $11.33^{\pm0.65}$ \\
MDM~\cite{mdm} & $15.96^{\pm1.92}$ & $\mathbf{1.000^{\pm0.0000}}$ & $\underline{5.98^{\pm0.15}}$ & $16.43^{\pm0.50}$ & $\underline{18.40^{\pm7.95}}$ & $\mathbf{0.647^{\pm0.0035}}$ & $5.89^{\pm0.33}$ & $10.96^{\pm0.27}$ \\
MDM-GRU~\cite{mdm} & $\underline{4.96^{\pm0.97}}$ & $0.995^{\pm0.0000}$ & $6.36^{\pm0.22}$ & $\underline{17.79^{\pm0.58}}$ & $18.63^{\pm25.87}$ & $0.574^{\pm0.0046}$ & $6.20^{\pm0.24}$ & $10.49^{\pm0.32}$ \\
\cellcolor{Gray}ReGenNet & \cellcolor{Gray}$\mathbf{0.27^{\pm0.03}}$ & \cellcolor{Gray}$\mathbf{1.000^{\pm0.0000}}$ & \cellcolor{Gray}$\mathbf{5.39^{\pm0.12}}$ & \cellcolor{Gray}$\mathbf{20.24^{\pm0.64}}$ & \cellcolor{Gray}$\mathbf{13.76^{\pm4.78}}$ & \cellcolor{Gray}$0.601^{\pm0.0040}$ & \cellcolor{Gray}$\underline{6.35^{\pm0.24}}$ & \cellcolor{Gray}$\mathbf{12.02^{\pm0.33}}$ \\
\bottomrule
\end{tabular}
}
\end{center}
\vspace{-4mm}
\caption[caption]{\textbf{Comparison to state-of-the-arts} on the \textit{online, unconstrained} setting for human action-reaction synthesis on Chi3D-AS. $\pm$ indicates 95\% confidence interval, $\rightarrow$ means that closer to Real is better. \textbf{Bold} indicates best result and \underline{underline} indicates second best.}
\label{tab:cmdm_chi3d}
\vspace{-2mm}
\end{table*}
  
\begin{table}[t]
\centering
\begin{center}
\resizebox{0.48\textwidth}{!}{
\begin{tabular}{ l c c c c c}
  \toprule
  \multirow{2}{1cm}{Methods}  & \multirow{2}{2.cm}{\centering R Precision (Top 3)$\uparrow$} & \multirow{2}{1.5cm}{\centering FID $\downarrow$} & \multirow{2}{2.5cm}{\centering MM Dist$\downarrow$}  & \multirow{2}{2cm}{\centering Diversity$\rightarrow $} & \multirow{2}{*}{\centering MModality $\uparrow$}\\
  \\
  \midrule
    Real & $0.722^{\pm0.004}$ & $0.002^{\pm0.0002}$ & $3.503^{\pm0.011}$ & $5.390^{\pm0.058}$ & - \\
    \midrule
    T2M~\cite{guo2022generating} & $0.224^{\pm0.003}$ & $32.482^{\pm0.0975}$ & $7.299^{\pm0.016}$ & $4.350^{\pm0.073}$ & $0.719^{\pm0.041}$ \\
    MDM~\cite{mdm} & $0.370^{\pm0.006}$ & $3.397^{\pm0.0352}$ & $8.640^{\pm0.065}$ & $4.780^{\pm0.117}$ & $2.288^{\pm0.039}$\\
    MDM-GRU~\cite{mdm} & $0.328^{\pm0.012}$ & $6.397^{\pm0.2140}$ & $8.884^{\pm0.040}$ & $\underline{4.851^{\pm0.081}}$ & $2.076^{\pm0.040}$\\
    RAIG~\cite{role_aware} & $0.363^{\pm0.008}$ & $\underline{2.915^{\pm0.0292}}$ & $\underline{7.294^{\pm0.027}}$ & $4.736^{\pm0.099}$ & $2.203^{\pm0.049}$\\
    InterGen~\cite{intergen} & $\underline{0.374^{\pm0.005}}$ & $13.237^{\pm0.0352}$ & $10.929^{\pm0.026}$ & $4.376^{\pm0.042}$ & $\mathbf{2.793^{\pm0.014}}$ \\
    \midrule
    \cellcolor{Gray}ReGenNet & \cellcolor{Gray}$\mathbf{0.407^{\pm0.003}}$ & \cellcolor{Gray}$\mathbf{2.265^{\pm0.0969}}$ & \cellcolor{Gray}$\mathbf{6.860^{\pm0.0.040}}$ & \cellcolor{Gray}$\mathbf{5.214^{\pm0.139}}$ & \cellcolor{Gray}$\underline{2.391^{\pm0.023}}$\\
  \bottomrule
\end{tabular}}
\end{center}
\vspace{-2mm}
\caption{\textbf{Comparison to state-of-the-arts} on the \textit{online, unconstrained} setting for human action-reaction synthesis on the InterHuman-AS dataset. $\pm$ indicates 95\% confidence interval, $\rightarrow$ means that closer to Real is better. \textbf{Bold} indicates best result and \underline{underline} indicates second best.}
\label{tab:cmdm_interhuman}
\end{table}

\begin{table}[t]
\vspace{-2mm}
\begin{center}
\resizebox{0.48\textwidth}{!}{
\begin{tabular}{ l c c c c}
\toprule
Method & FID$\downarrow$ & Acc.$\uparrow$ & Div.$\rightarrow$ & Multimod.$\rightarrow$\\
\midrule
Real & $0.10^{\pm0.00}$ & $0.849^{\pm0.0002}$ & $12.98^{\pm0.11}$ & $22.77^{\pm0.35}$ \\
\midrule
cVAE~\cite{vae} & $63.23^{\pm7.74}$ & $\underline{0.708^{\pm0.0004}}$ & $11.15^{\pm0.03}$ & $17.34^{\pm0.23}$ \\
AGRoL~\cite{agrol} & $35.83^{\pm1.13}$ & $0.592^{\pm0.0003}$ & $\underline{12.42^{\pm0.06}}$ & $18.67^{\pm0.21}$ \\
MDM~\cite{mdm} & $36.75^{\pm2.87}$ & $0.692^{\pm0.0004}$ & $11.73^{\pm0.05}$ & $18.24^{\pm0.21}$ \\
MDM-GRU~\cite{mdm} & $\underline{25.57^{\pm1.71}}$ & $0.636^{\pm0.0005}$ & $\mathbf{13.20^{\pm0.09}}$ & $\underline{20.49^{\pm0.33}}$ \\
\cellcolor{Gray}ReGenNet & \cellcolor{Gray}$\mathbf{8.16^{\pm0.42}}$ & \cellcolor{Gray}$\mathbf{0.713^{\pm0.0002}}$ & \cellcolor{Gray}$13.88^{\pm0.13}$ & \cellcolor{Gray}$\mathbf{21.63^{\pm0.41}}$ \\
\bottomrule
\end{tabular}}
\end{center}
\vspace{-4mm}
\caption[caption]{\textbf{Generalization results} to different viewpoints on the \textit{online, unconstrained} setting on the NTU120-AS dataset. \textbf{Bold} indicates best result and \underline{underline} indicates second best.}
\label{tab:camera2}
\end{table}
    
\noindent\textbf{Explicit interaction loss.}
Inspired by previous generative models of human scene/object interaction counterparts, which designed delicate distance-based representations to model interactions, we design explicit interaction losses to measure the relative distances of the interacted spatiotemporal body pose $\bm{\theta}$, orientation $\bm{q}$ and translation $\bm{\gamma}$ as
\begin{equation} \label{eqn:rep}
  \begin{aligned}
  \bm{\theta}^{x\to y} &= FK(\bm{\theta}^y) - FK(\bm{\theta}^x); \\ \bm{q}^{x\to y} &= RM(\bm{q}^y)^\top\cdot RM(\bm{q}^x); \\ \bm{\gamma}^{x\to y} &= \bm{\gamma}^y - \bm{\gamma}^x,
  \end{aligned}
\end{equation}
where $FK(\cdot)$ denotes the forward kinematic function to convert the rotation pose into joint positions, and $RM(\cdot)$ converts the rotation poses to rotation matrices. The interaction loss is formulated as
\begin{equation} \label{eqn:loss_interact}
  \begin{aligned}
  \mathcal{L}_{inter} &= \frac{1}{N}\biggl(\sum_{i =1}^{N} \| \bm{\theta}^{x_0\to y} - \bm{\theta}^{\hat{x}_0\to y}\|_{2}^{2} \\ &+ \sum_{i =1}^{N} \| \bm{q}^{x_0\to y} - \bm{q}^{\hat{x}_0\to y}\|_{2}^{2} + \sum_{i =1}^{N} \| \bm{\gamma}^{x_0\to y} - \bm{\gamma}^{\hat{x}_0\to y}\|_{2}^{2}\biggr).
  \end{aligned}
\end{equation}
Overall, the training loss is $\mathcal{L}_{all} = \mathcal{L}_{dm} + \lambda_{inter}\cdot\mathcal{L}_{inter}$, and $\lambda_{inter}$ is the loss weight. 

\noindent\textbf{Customization Support.} We tackle the most challenging setting of \textit{online} action-reaction synthesis without seeing the future motions, even being agnostic to the actor's intentions. There exist other scenarios that relax these restrictions, such as \textit{offline} generative models if the time delay can be tolerated. Our proposed \textbf{ReGenNet} is modular, flexible, and can be trimmed for other practical usages. 1) The intention branch can be enabled if the actor's intention is available to the reactor, or removed otherwise; 2) The directional attention mask can be disabled for \textit{offline} models.

\section{Experiment}

First, we give the definitions of our experiment settings. We name the setting of instant human action-reaction synthesis without seeing the future motions as \textbf{online} models, and the opposite is \textbf{offline} models to relax the synchronicity requirements. We also define \textbf{unconstrained} and \textbf{constrained} settings depending on whether the intention of the actor is visible to the reactor. We mainly focus on the challenging \textbf{online, unconstrained} setting due to its potential for practical applications.

\begin{table*}[htbp]
  \vspace{-1mm}
  \begin{center}
  \resizebox{1.0\textwidth}{!}{
  \begin{tabular}{l l c c c c c c c c}
  \toprule
  \multirow{2}{*}{Class} & \multirow{2}{*}{Settings} & \multicolumn{4}{c}{Train conditioned} & \multicolumn{4}{c}{Test conditioned} \\
  \cmidrule(rl){3-6} \cmidrule(rl){7-10}
  & & FID$\downarrow$ & Acc.$\uparrow$ & Div.$\rightarrow$ & Multimod.$\rightarrow$ & FID$\downarrow$ & Acc.$\uparrow$ & Div.$\rightarrow$ & Multimod.$\rightarrow$ \\
  \midrule
  & Real & $0.094^{\pm0.0003}$ & $1.000^{\pm0.00}$ & $10.542^{\pm0.0632}$ & $26.709^{\pm0.6193}$ & $0.085^{\pm0.0003}$ & $0.867^{\pm0.0002}$ & $13.063^{\pm0.0908}$ & $25.032^{\pm0.2332}$ \\
  \midrule
  \multirow{2}{*}{Modules} & 1) Add & $0.975^{\pm0.0186}$ & $1.000^{\pm0.00}$ & $10.685^{\pm0.0493}$ & $26.272^{\pm0.3663}$ & $12.828^{\pm0.9335}$ & $0.747^{\pm0.0003}$ & $13.721^{\pm0.1513}$ & $22.771^{\pm0.1921}$\\
  &2) w.o. PE & $0.892^{\pm0.0130}$ & $1.000^{\pm0.00}$ & $10.717^{\pm0.0567}$ & $26.345^{\pm0.3432}$ & $12.916^{\pm1.2802}$ & $0.747^{\pm0.0001}$ & $13.775^{\pm0.1526}$ & $22.752^{\pm0.1330}$ \\
  \midrule
  \multirow{1}{*}{Loss} & w.o. $\mathcal{L}_{inter}$ & $1.960^{\pm0.0621}$ & $0.999^{\pm0.00}$ & $10.778^{\pm0.0597}$ & $26.024^{\pm0.3223}$ & $12.146^{\pm0.9044}$ & $0.751^{\pm0.0002}$ & $13.727^{\pm0.1808}$ & $22.846^{\pm0.1606}$ \\
  \midrule
  \multirow{4}{*}{Num. of $\ell_{dec}$} & 2 & $11.445^{\pm0.8738}$ & $0.993^{\pm0.00}$ & $10.972^{\pm0.0543}$ & $24.815^{\pm0.3769}$ & $29.015^{\pm3.7751}$ & $0.744^{\pm0.0002}$ & $13.107^{\pm0.1273}$ & $21.134^{\pm0.1150}$ \\
  & 4 & $3.218^{\pm0.1120}$ & $0.999^{\pm0.00}$ & $10.856^{\pm0.0521}$ & $25.728^{\pm0.3790}$ & $18.148^{\pm1.7413}$ & $0.743^{\pm0.0002}$ & $13.418^{\pm0.1048}$ & $21.813^{\pm0.1671}$ \\
  & 12 & $1.967^{\pm0.0287}$ & $1.000^{\pm0.00}$ & $10.752^{\pm0.0533}$ & $26.038^{\pm0.3370}$ & $13.348^{\pm0.7420}$ & $0.725^{\pm0.0002}$ & $14.090^{\pm0.1629}$ & $22.906^{\pm0.1197}$ \\
  & 16 & $2.382^{\pm0.0511}$ & $1.000^{\pm0.00}$ & $10.757^{\pm0.0509}$ & $25.908^{\pm0.3370}$ & $11.089^{\pm1.3902}$ & $0.728^{\pm0.0002}$ & $14.173^{\pm0.1327}$ & $23.069^{\pm0.2492}$ \\
  \midrule
  \cellcolor{Gray}Final Version & \cellcolor{Gray}ReGenNet & \cellcolor{Gray}$0.896^{\pm0.0109}$ & \cellcolor{Gray}$1.000^{\pm0.00}$ & \cellcolor{Gray}$10.694^{\pm0.0542}$ & \cellcolor{Gray}$26.247^{\pm0.3476}$ & \cellcolor{Gray}$10.999^{\pm0.7425}$ & \cellcolor{Gray}$0.749^{\pm0.0002}$ & \cellcolor{Gray}$13.797^{\pm0.1593}$ & \cellcolor{Gray}$22.902^{\pm0.1353}$ \\
  \bottomrule
  \end{tabular}
  }
  \end{center}
  \vspace{-4mm}
  \caption[caption]{\textbf{Ablation studies} on the \textit{online, unconstrained} setting on the NTU120-AS dataset.}
  \label{tab:ablations}
  \end{table*}

\begin{table*}[t]
  \vspace{-2mm}
  \begin{center}
  \resizebox{1.0\textwidth}{!}{
  \begin{tabular}{c c c c c c c c c c}
  \toprule
  \multirow{2}{*}{Timesteps} & \multicolumn{4}{c}{Train conditioned} & \multicolumn{4}{c}{Test conditioned} & \multirow{2}{*}{Latency(ms)} \\
  \cmidrule(rl){2-5} \cmidrule(rl){6-9}
  & FID$\downarrow$ & Acc.$\uparrow$ & Div.$\rightarrow$ & Multimod.$\rightarrow$ & FID$\downarrow$ & Acc.$\uparrow$ & Div.$\rightarrow$ & Multimod.$\rightarrow$ & \\
  \midrule
  Real & $0.094^{\pm0.0003}$ & $1.000^{\pm0.00}$ & $10.542^{\pm0.0632}$ & $26.709^{\pm0.6193}$ & $0.085^{\pm0.0003}$ & $0.867^{\pm0.0002}$ & $13.063^{\pm0.0908}$ & $25.032^{\pm0.2332}$ & - \\
  \midrule
  2 & $0.912^{\pm0.0110}$ & $1.000^{\pm0.00}$ & $10.688^{\pm0.0529}$ & $26.249^{\pm0.3504}$ & $12.132^{\pm0.8301}$ & $0.751^{\pm0.0002}$ & $13.707^{\pm0.1683}$ & $22.797^{\pm0.1172}$ & $0.33$\\
  \cellcolor{Gray}5 & \cellcolor{Gray}$0.896^{\pm0.0109}$ & \cellcolor{Gray}$1.000^{\pm0.00}$ & \cellcolor{Gray}$10.694^{\pm0.0542}$ & \cellcolor{Gray}$26.247^{\pm0.3476}$ & \cellcolor{Gray}$10.999^{\pm0.7425}$ & \cellcolor{Gray}$0.749^{\pm0.0002}$ & \cellcolor{Gray}$13.797^{\pm0.1593}$ & \cellcolor{Gray}$22.902^{\pm0.1353}$ & $\cellcolor{Gray}0.76$\\
  10 & $0.903^{\pm0.0108}$ & $1.000^{\pm0.00}$ & $10.691^{\pm0.0536}$ & $26.250^{\pm0.3507}$ & $11.466^{\pm0.8199}$ & $0.749^{\pm0.0002}$ & $13.762^{\pm0.1647}$ & $22.855^{\pm0.1264}$ & $1.58$ \\
  100 & $0.897^{\pm0.0109}$ & $1.000^{\pm0.00}$ & $10.692^{\pm0.0537}$ & $26.248^{\pm0.3491}$ & $11.082^{\pm0.7440}$ & $0.748^{\pm0.0002}$ & $13.794^{\pm0.1581}$ & $22.890^{\pm0.1223}$ & $15.17$\\
  1000 & $0.908^{\pm0.0109}$ & $1.000^{\pm0.00}$ & $10.692^{\pm0.0540}$ & $26.247^{\pm0.3502}$ & $11.719^{\pm0.8059}$ & $0.750^{\pm0.0001}$ & $13.738^{\pm0.1665}$ & $22.830^{\pm0.1220}$ & $150.69$\\
  \bottomrule
  \end{tabular}
  }
  \end{center}
  \vspace{-4mm}
  \caption[caption]{\textbf{Ablation Studies} of the number of DDIM~\cite{ddim} sampling timesteps on the \textit{online, unconstrained} setting on NTU120-AS.}
  \label{tab:ddim_step}
  \end{table*}

\begin{table}[t]
  \vspace{-2mm}
  \begin{center}
  \resizebox{0.48\textwidth}{!}{
  \begin{tabular}{ l c c c c}
  \toprule
  Method & FID$\downarrow$ & Acc.$\uparrow$ & Div.$\rightarrow$ & Multimod.$\rightarrow$\\
  \midrule
  Real & $0.09^{\pm0.00}$ & $0.867^{\pm0.0002}$ & $13.06^{\pm0.09}$ & $25.03^{\pm0.23}$ \\
  \midrule
  Random & $12.69^{\pm1.22}$ & $0.656^{\pm0.0002}$ & $13.97^{\pm0.08}$ & $22.19^{\pm0.34}$ \\
  \cellcolor{Gray}ReGenNet & \cellcolor{Gray}$\mathbf{11.00^{\pm0.74}}$ & \cellcolor{Gray}$\mathbf{0.749^{\pm0.0002}}$ & \cellcolor{Gray}$\mathbf{13.80^{\pm0.16}}$ & \cellcolor{Gray}$\mathbf{22.90^{\pm0.14}}$ \\
  \bottomrule
  \end{tabular}}
  \end{center}
  \vspace{-4mm}
  \caption[caption]{\textbf{Ablation studies} on the necessity of the explicit actor-reactor order annotations on the NTU120-AS dataset.}
  \label{tab:need_label}
  \end{table}

\subsection{Datasets and Evaluation Metrics}

We evaluate our model on our proposed NTU120-AS, InterHuman-AS and Chi3D-AS datasets with SMPL-X~\cite{smplx} body models and actor-reactor annotations. NTU120-AS includes 8,118 human interaction sequences captured by 3 cameras of 26 action categories. We choose camera 1 and follow the cross-subject protocol where half of the subjects are used for training and the remaining ones for testing. We also evaluate our model on camera 2 to examine the generalization ability for viewpoint changes. InterHuman-AS presents 6,022 interaction sequences captured by multi-view motion capture studio. Chi3D-AS contains 373 interaction sequences and we randomly split the training and test sets with a ratio of 4:1. In all our experiments, we adopt the 6D rotation representation~\cite{6d_rot} as in~\cite{actor}.

For evaluation metrics, we follow prior works in human motion generation~\cite{action2motion,actor,mdm} and use Frechet Inception Distance (FID)~\cite{fid}, action recognition accuracy, diversity, and multi-modality for quantitative evaluations. For all these metrics, we train the action recognition model~\cite{stgcn} to extract motion features for calculating FID, diversity, and multi-modality, or directly compute the action recognition accuracy. Following~\cite{actformer}, the root translation is considered for the action recognition model since relative root translations matter for person-person interactions.
We generate the reactions by sampling actor motions from the training or test sets as \textbf{train-conditioned} and \textbf{test-conditioned}, respectively. Evaluation results for test conditioned examine the capability of reaction generation for unseen actor motions.
We generate 1,000 samples 20 times with different random seeds and report the average with the confidence interval at 95\% as previous works~\cite{action2motion,actor,mdm}.

\subsection{Implementation Details} We train our diffusion-based model with $T=1,000$ noising timesteps and a cosine noise schedule in a classifier-free manner~\cite{ho2022classifier}. The batch size is set as 64 for NTU120-AS, InterHuman-AS and 16 for Chi3D-AS, the number of decoder layers is 8 and the latent dimension of the Transformer tokens is 512. For the loss weights, we set $\lambda_{inter}=1$.
Each model is trained for 500K steps on 4 NVIDIA A100 GPUs within 20 hours.
During inference, we adopt the DDIM~\cite{ddim} strategy to accelerate. Unless specified, we run 5 DDIM sampling steps for all the diffusion-based models in our experiments.

\begin{table*}[hbtp]
  \vspace{-1mm}
  \begin{center}
  \resizebox{1.0\textwidth}{!}{
  \begin{tabular}{l c c c c c c c c}
  \toprule
  \multirow{2}{*}{Method} & \multicolumn{4}{c}{Train conditioned} & \multicolumn{4}{c}{Test conditioned}\\
  \cmidrule(rl){2-5} \cmidrule(rl){6-9}
  & FID$\downarrow$ & Acc.$\uparrow$ & Div.$\rightarrow$ & Multimod.$\rightarrow$ & FID$\downarrow$ & Acc.$\uparrow$ & Div.$\rightarrow$ & Multimod.$\rightarrow$ \\
  \midrule
  Real & $0.09^{\pm0.00}$ & $1.000^{\pm0.0000}$ & $10.54^{\pm0.06}$ & $26.71^{\pm0.62}$ & $0.09^{\pm0.00}$ & $0.867^{\pm0.0002}$ & $13.06^{\pm0.09}$ & $25.03^{\pm0.23}$ \\
  \midrule
  cVAE~\cite{vae} & $81.62^{\pm14.43}$ & $0.902^{\pm0.0002}$ & $10.10^{\pm0.02}$ & $19.38^{\pm0.16}$ & $74.73^{\pm4.86}$ & $0.760^{\pm0.0002}$ & $11.14^{\pm0.04}$ & $18.40^{\pm0.26}$ \\
  AGRoL~\cite{agrol} & $10.87^{\pm1.07}$ & $0.983^{\pm0.0000}$ & $11.45^{\pm0.07}$ & $23.80^{\pm0.42}$ & $16.55^{\pm1.41}$ & $0.716^{\pm0.0002}$ & $13.84^{\pm0.10}$ & $21.73^{\pm0.20}$ \\
  MDM~\cite{mdm} & $\underline{1.61^{\pm0.02}}$ & $\underline{0.999^{\pm0.0000}}$ & $\underline{10.76^{\pm0.06}}$ & $\underline{26.02^{\pm0.30}}$ & $\underline{7.49^{\pm0.62}}$ & $\mathbf{0.775^{\pm0.0003}}$ & $\underline{13.67^{\pm0.18}}$ & $\underline{24.14^{\pm0.29}}$ \\
  MDM-GRU~\cite{mdm} & $5.31^{\pm0.18}$ & $0.993^{\pm0.0000}$ & $11.03^{\pm0.06}$ & $25.04^{\pm0.36}$ & $24.25^{\pm1.39}$ & $0.720^{\pm0.0002}$ & $\mathbf{13.43^{\pm0.09}}$ & $22.24^{\pm0.29}$ \\
  \cellcolor{Gray}ReGenNet & \cellcolor{Gray}$\mathbf{0.64^{\pm0.01}}$ & \cellcolor{Gray}$\mathbf{1.000^{\pm0.0000}}$ & \cellcolor{Gray}$\mathbf{10.70^{\pm0.05}}$ & \cellcolor{Gray}$\mathbf{26.36^{\pm0.38}}$ & \cellcolor{Gray}$\mathbf{6.19^{\pm0.33}}$ & \cellcolor{Gray}$\underline{0.772^{\pm0.0003}}$ & \cellcolor{Gray}$14.03^{\pm0.09}$ & \cellcolor{Gray}$\mathbf{25.21^{\pm0.34}}$ \\
  \bottomrule
  \end{tabular}
  }
  \caption[caption]{\textbf{Results} on the \textbf{offline}\textit{, unconstrained} setting on NTU120-AS. \textbf{Bold} indicates best result and \underline{underline} indicates second best.}
  \label{tab:cmdm_offline}
  \end{center}
  \end{table*}
  
  \begin{table*}[hbtp]
  \vspace{-2mm}
  \begin{center}
  \resizebox{1.0\textwidth}{!}{
  \begin{tabular}{l c c c c c c c c}
  \toprule
  \multirow{2}{*}{Method} & \multicolumn{4}{c}{Train-conditioned} & \multicolumn{4}{c}{Test-conditioned}\\
  \cmidrule(rl){2-5} \cmidrule(rl){6-9}
  & FID$\downarrow$ & Acc.$\uparrow$ & Div.$\rightarrow$ & Multimod.$\rightarrow$ & FID$\downarrow$ & Acc.$\uparrow$ & Div.$\rightarrow$ & Multimod.$\rightarrow$ \\
  \midrule
  Real & $0.09^{\pm0.00}$ & $1.000^{\pm0.0000}$ & $10.54^{\pm0.06}$ & $26.71^{\pm0.62}$ & $0.09^{\pm0.00}$ & $0.867^{\pm0.0002}$ & $13.06^{\pm0.09}$ & $25.03^{\pm0.23}$ \\
  \midrule
  ReGenNet-unconstrained & {$\underline{0.90^{\pm0.01}}$} & {$\mathbf{1.000^{\pm0.0000}}$} & $\mathbf{10.69^{\pm0.05}}$ & $\underline{26.25^{\pm0.35}}$ & $\underline{11.00^{\pm0.74}}$ & $\underline{0.749^{\pm0.0002}}$ & $\mathbf{13.80^{\pm0.16}}$ & $\underline{22.90^{\pm0.14}}$ \\
  \midrule
  \cellcolor{Gray}ReGenNet-constrained & \cellcolor{Gray}$\mathbf{0.86^{\pm0.01}}$ & \cellcolor{Gray}$\mathbf{1.000^{\pm0.0000}}$ & \cellcolor{Gray}$\underline{10.69^{\pm0.06}}$ & \cellcolor{Gray}$\mathbf{26.26^{\pm0.35}}$ & \cellcolor{Gray}$\mathbf{5.89^{\pm0.24}}$ & \cellcolor{Gray}$\mathbf{0.902^{\pm0.0001}}$ & \cellcolor{Gray}$\underline{11.90^{\pm0.06}}$ & \cellcolor{Gray}$\mathbf{25.08^{\pm0.20}}$ \\
  \bottomrule
  \end{tabular}
  }
  \end{center}
  \vspace{-4mm}
  \caption[caption]{\textbf{Results} on the \textit{online, }\textbf{constrained} setting on NTU120-AS. \textbf{Bold} indicates best result and \underline{underline} indicates second best.}
  \label{tab:cmdm_ac}
\end{table*}

\subsection{Comparison to State-of-the-arts}

In this section, we evaluate our model on the most challenging \textbf{online, unconstrained} setting, \ieno, to generate instant reactions without knowing the intention of the actors.
Since there is no previous work tackling the conditional human action-reaction synthesis setting, we adopt and re-implement some previous works for human-centric generative models as baselines 1) cVAE~\cite{vae}, which is widely adopted in previous human-scene/object interaction generative models; 2) MDM~\cite{mdm}, the state-of-the-art diffusion-based method for human motion generation and MDM-GRU~\cite{mdm}, a diffusion-based model with a GRU~\cite{cho2014learning} backbone; 3) AGRoL~\cite{agrol}, the state-of-the-art method to generate full-body motions from sparse tracking signals, implemented by diffusion models with MLPs as backbones. At the inference stage, we employ a sliding-window strategy to iteratively generate the reactions for the \textbf{online} setting. For a fair comparison, we also run 5 DDIM sampling steps for AGRoL~\cite{agrol}, MDM~\cite{mdm} and MDM-GRU~\cite{mdm} baselines. For the text-conditioned setting, we adopt T2M~\cite{guo2022generating}, MDM~\cite{mdm}, MDM-GRU~\cite{mdm}, RAIG~\cite{role_aware} and InterGen~\cite{intergen} as baselines.

\cref{tab:cmdm},~\cref{tab:cmdm_chi3d} and show the comparisons on the NTU120-AS and Chi3D-AS dataset, respectively.
For the two datasets, our proposed \textbf{ReGenNet} notably outperforms baselines on the FID metric, which shows that our generation is closest to the real human reaction distributions.
For the train-conditioned setting where the actor motions are sampled from the training set, our method yields state-of-the-art performance for FID, action recognition accuracy, diversity, and multi-modality on NTU120-AS and Chi3D-AS datasets (except for second best for the diversity of NTU120-AS), which shows that our model learns the reaction patterns well. For the test-conditioned setting, our method achieves the best FID, action recognition accuracy, and multi-modality for the NTU120-AS dataset and the best FID, and multi-modality for the Chi3D-AS dataset. This verifies that our method can generalize well to unseen actor motions. Due to the limited scale of the Chi3D-AS test set, there might be some fluctuations in the experimental results.
~\cref{tab:cmdm_interhuman} shows the comparisons on the InterHuman-AS dataset, our method also yields the best results compared to the baselines.

\subsection{Generalization Experiments}

To verify the generalization ability of our model to viewpoint changes, we train the generative models on camera 1 and inference on camera 2 of the NTU120-AS dataset. As reported in~\cref{tab:camera2}, \textbf{ReGenNet} achieves the best FID score, action recognition accuracy, and multi-modality than other baselines, which shows that our method is robust to viewpoint changes.

\subsection{Ablation Study}

\noindent{\textbf{Basic Module Designs.}}
As illustrated in~\cref{sec:regennet}, a simple yet effective conditioning scheme is to concatenate the actor and reactor motion features as the input to the Transformer decoders. We also tried to add these features together and the results are listed on the ``Modules - 1) Add'' setting in~\cref{tab:ablations}. However, the results become inferior in terms of the FID and action recognition accuracy metrics. We also verify that adding positional embedding is effective to bring lower FID scores for the test-conditioned setting.

\noindent{\textbf{Explicit Interaction Loss.}}
We design distance-based explicit interaction loss based on the relative interacted body pose, head orientation, and positions as~\cref{eqn:loss_interact}. From the ``Loss - w.o $\mathcal{L}_{inter}$'' setting in~\cref{tab:ablations}, we derive that the explicit interaction loss contributes to lower FIDs with minor action recognition accuracy drop for the test conditioned setting ($0.751 \to 0.749$). This confirms that the explicit interaction loss enhances the reaction generation quality.

\noindent{\textbf{Layer of Decoders.}}
We set the number of Transformer decoder units layers $\ell_{dec}=8$ in our ultimate model, and we also ablate $\ell_{dec}=2, 4, 12, 16$. As represented on the ``Num. of $\ell_{dec}$'' setting of~\cref{tab:ablations}, we can observe that setting $\ell_{dec}=8$ obtains the lowest FID score and best overall performance.

\begin{figure*}[h]
  \centering
  \includegraphics[width=1.0\linewidth]{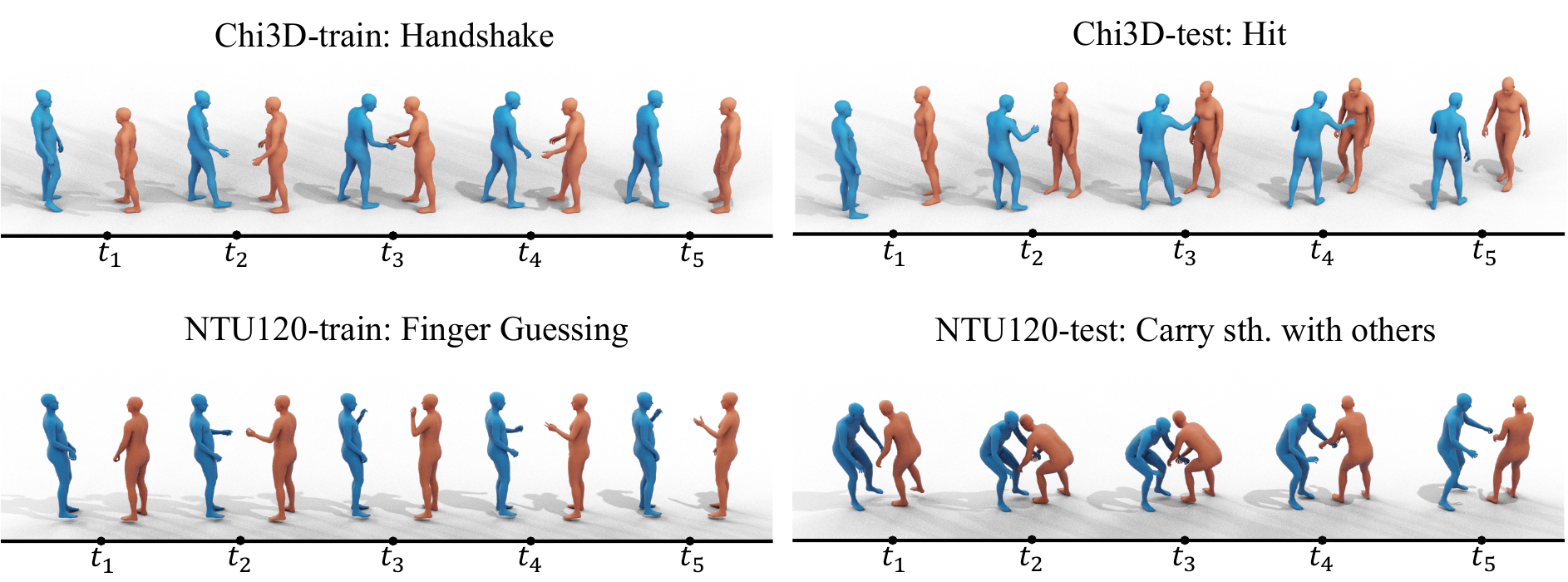}
  \vspace{-4mm}
  \caption{\textbf{Visualization} of human action-reaction synthesis results. \textcolor{blue}{Blue} for actors and \textcolor{orange}{Orange} for reactors.}
  \label{fig:vis}
  \vspace{-2mm}
\end{figure*}

\noindent{\textbf{Number of DDIM sampling timesteps.}}
We report the reaction generation results and the latency for per frame reaction generation under different DDIM~\cite{ddim} sampling timesteps, \ieno, 2, 5, 10, 10, 100, and 1000. We take our trained \textbf{ReGenNet} with 1,000 sampling timesteps and inference with a subset of diffusion steps. From~\cref{tab:ddim_step}, we can see that 5 DDIM timesteps inference yields the best FID score with low latency. Thus, we adopt the DDIM-5 inference for all the reported results.

\noindent{\textbf{Necessity of actor-reactor annotations.}}
To verify the necessity of explicitly annotating the actor-reactor orders, we ablate it by randomly shuffling the actor-reactor roles as an unsupervised way for human reaction generation. The results depicted in~\cref{tab:need_label} show that random shuffling brings inferior performance than ours. One possible explanation could be that the action pattern and reaction pattern differ a lot, yet directly randomly shuffling the actor-reactor order may distract the reaction generation model.

\subsection{Extension to other settings}

Our proposed \textbf{ReGenNet} is modular and can be trimmed for other scenarios. We show the experimental results of our model adapting to \textbf{offline} (see~\cref{tab:cmdm_offline}) and \textbf{constrained} (see~\cref{tab:cmdm_ac}) settings. For the \textbf{offline} setting, we replace the Transformer decoder units equipped with attention masks with an 8-layer Transformer encoder architecture. As shown in~\cref{tab:cmdm_offline}, our model achieves superior performance on most of the metrics, which shows the effectiveness of our method and flexibility for adaptation. For the \textbf{constrained} setting where the actor's intention is available to the reactor, we embed the action label $a$ into \textbf{ReGenNet} as described in~\cref{sec:regennet}. As expected, the constrained setting achieves superior performance than the unconstrained setting since the action serves as a strong hint to generate the reactions.

\subsection{Qualitative evaluation.}

We visualize some human reaction generation examples in~\cref{fig:vis}, sampled from the train/test sets of Chi3D-AS and NTU120-AS datasets. The visualization results show that \textbf{ReGenNet} can synthesize human reactions with plausible 1) relative head orientations, \ieno, handshaking face-to-face; 2) position change, \ieno, step back for hitting; 3) body part movements and hand gestures, \ieno, realistic hand pose; and 4) semantics, \ieno, the action between the actor and reactor is visually reasonable. For more visualizations and \textbf{videos} of the generated human reactions, please refer to the supplementary materials.

\section{Conclusion and Limitations}

In this paper, we propose the \textit{first} multi-setting human action-reaction synthesis benchmark with a comprehensive analysis of the asymmetric, dynamic, synchronous, and detailed characteristics of human-human interactions. For the first time, we annotate the actor-reactor order for the NTU120, Chi3D, and InterHuman datasets. We propose \textbf{ReGenNet}, a conditional diffusion model with a Transformer decoder architecture combined with an explicit distance-based interaction loss.
Extensive experiments demonstrated that ReGenNet can generate instant and realistic reactions, even being agnostic to the actor's intentions. We also verify that our method is generalizable to viewpoint changes. Furthermore, experimental results show that ReGenNet is modular and can be trimmed for different settings of conditional action-reaction generations.

\noindent{\textbf{Limitations.}}
The setup of our benchmark and datasets still has some limitations: 1) \textbf{Setup:} Real-world human-human interactions are much more complicated with longer durations, interaction patterns, and actor-reactor transitions. Currently, we only focus on the human action-reaction synthesis within atomic action periods, which could be improved in the future; 2) \textbf{Datasets:} 
The human motion of the NTU120 dataset extracted by algorithms is noisy, even with human-human inter-penetrations. The facial expressions for these datasets are also not natural. Therefore, high-quality human-human interaction datasets with actor-reactor annotations are desired in the future.

\noindent\textbf{Acknowledgments.} 
This work was supported in part by NSFC (62201342, 62101325), Shanghai Municipal Science and Technology Major Project (2021SHZDZX0102), NSFC under Grant 62302246 and ZJNSFC under Grant LQ23F010008.
This work is supported by High Performance Computing Center at Eastern Institute of Technology, Ningbo.

{
    \small
    \bibliographystyle{ieeenat_fullname}
    \bibliography{main}
}

\appendix

\begingroup

\appendix
\twocolumn[
\begin{center}
\Large{\bf ReGenNet: Towards Human Action-Reaction Synthesis \\ **Appendix**}
\end{center}
]

\counterwithin{table}{section}
\counterwithin{figure}{section}
\setcounter{page}{1}

\section{Details of building the datasets}
\label{append:dataset}
In this section, we provide more details of the definitions of asymmetric actions and the actor-reactor order labeling process. Taking the Chi3D-AS dataset as an example, it contains 8 action categories, \ieno, ``Grab'', ``Handshake'', ``Hit'', ``HoldingHands'', ``Hug'', ``Kick'', ``Posing'' and ``Push''. For all these actions, there always exists a person acts and the other reacts. Even for the interaction of ``Handshake'', a man actively reaches out his hands and the other one consequently reaches out his hands to complete the handshaking.

To annotate the actor-reactor order of the NTU120 and Chi3D datasets, we first build an annotation tool based on Python3, tkinter and tkVideoPlayer. Then we render the motion sequences of NTU120 and Chi3D to videos. The annotation tool displays the motion sequence videos together with the corresponding RGB videos. The annotators select the actor-reactor order of the given two-person interactions based on both the motion and the RGB videos. After the initial annotation process, the results are double-checked manually. The failed extracted motion sequences by PyMAF-X~\cite{pymaf-x} for NTU120 are removed from the final datasets.

\section{The choice of the datasets}

We choose Chi3D~\cite{chi3d} for its ground-truth SMPL-X parameters with subtle hand gestures, which is suitable for the \textbf{detailed} feature of human-human interactions. The NTU120 dataset~\cite{nturgbd120} is chosen for that it is currently the largest human-human interaction dataset with 26 action categories and diverse reaction patterns. Take the action ``kick'' as an example, the reaction could be ``move back'', ``raise the feet'', ``jump backward'' or ``turn the shoulders''. Moreover, NTU-X~\cite{ntu-x} also verifies that it is feasible to extract SMPL-X parameters for NTU120, yet NTU-X is not open-source. The InterHuman dataset~\cite{intergen} is suitable for our setting, while it only contains human body parameters without dexterous hand movements.

\section{Effects of pose estimation results}

To \textit{quantitatively measure the effects brought by the noise of the pose estimation models}, we adopt the PyMAF-X to extract the SMPL-X parameters from the pure RGB videos of the \textbf{Chi3D dataset} and then use the estimated pose results for reaction generation. The experimental results in~\cref{tab:choose_ntu} show consistent improvements of our proposed ReGenNet over the baselines.

\begin{table*}[t]
  \begin{center}
  \resizebox{1.0\textwidth}{!}{
  \begin{tabular}{l c c c c c c c c }
  \toprule
  \multirow{2}{*}{Method} & \multicolumn{4}{c}{Train conditioned} & \multicolumn{4}{c}{Test conditioned} \\
  \cmidrule(rl){2-5} \cmidrule(rl){6-9}
  & FID$\downarrow$ & Acc.$\uparrow$ & Div.$\rightarrow$ & Multimod.$\rightarrow$ & FID$\downarrow$ & Acc.$\uparrow$ & Div.$\rightarrow$ & Multimod.$\rightarrow$ \\
  \midrule
  Real & $0.17^{\pm0.01}$ & $1.000^{\pm0.0000}$ & $5.28^{\pm0.11}$ & $20.26^{\pm0.65}$ & $1.06^{\pm1.94}$ & $0.570^{\pm0.0056}$ & $7.37^{\pm1.17}$ & $12.38^{\pm1.33}$ \\
  \midrule
  cVAE & $\underline{22.04^{\pm3.85}}$ & $0.800^{\pm0.0008}$ & $9.08^{\pm0.47}$ & $13.22^{\pm0.16}$ & $\mathbf{19.95^{\pm19.37}}$ & $0.374^{\pm0.0044}$ & $8.78^{\pm0.61}$ & $\mathbf{11.38^{\pm0.75}}$ \\
  AGRoL & $36.54^{\pm2.75}$ & $0.937^{\pm0.0001}$ & $7.02^{\pm0.16}$ & $14.66^{\pm0.24}$ & $\underline{21.09^{\pm24.31}}$ & $0.370^{\pm0.0058}$ & $\underline{6.68^{\pm1.09}}$ & $10.49^{\pm1.82}$ \\
  MDM & $22.92^{\pm1.29}$ & $\underline{0.979^{\pm0.0000}}$ & $6.34^{\pm0.14}$ & $\underline{15.76^{\pm0.41}}$ & $26.21^{\pm14.11}$ & $0.364^{\pm0.0053}$ & $6.49^{\pm0.50}$ & $9.83^{\pm1.01}$ \\
  MDM-GRU & $24.86^{\pm1.72}$ & $0.908^{\pm0.0000}$ & $\underline{6.23^{\pm0.12}}$ & $15.37^{\pm0.36}$ & $25.60^{\pm11.58}$ & $\underline{0.392^{\pm0.0066}}$ & $6.33^{\pm0.54}$ & $9.59^{\pm1.01}$ \\
  \cellcolor{Gray}ReGenNet & \cellcolor{Gray}$\mathbf{0.34^{\pm0.01}}$ & \cellcolor{Gray}$\mathbf{1.000^{\pm0.0000}}$ & \cellcolor{Gray}$\mathbf{5.52^{\pm0.11}}$ & \cellcolor{Gray}$\mathbf{20.06^{\pm0.48}}$ & \cellcolor{Gray}$22.88^{\pm14.12}$ & \cellcolor{Gray}$\mathbf{0.414^{\pm0.0064}}$ & \cellcolor{Gray}$\mathbf{7.03^{\pm0.79}}$ & \cellcolor{Gray}$\underline{10.75^{\pm1.37}}$ \\
  \bottomrule
  \end{tabular}
  }
  \end{center}
  \vspace{-4mm}
  \caption[caption]{Results on the Chi3D dataset with estimated pose results. $\pm$ indicates 95\% confidence interval, $\rightarrow$ means that closer to Real is better. \textbf{Bold} indicates best result and \underline{underline} indicates second best.}
  \label{tab:choose_ntu}
\end{table*}

\section{The technical details of the baselines}

For the Chi3D-AS and NTU120-AS datasets, we choose cVAE~\cite{vae}, AGRoL~\cite{agrol}, MDM~\cite{mdm}, and MDM-GRU~\cite{mdm} as our baselines for that they are state-of-the-art methods for human scene/object interaction generation and conditional human motion generation. We adopt the 6D rotation representation of the SMPL-X parameters as inputs and the frame length is set to 60 and 150 for NTU120-AS and Chi3d-AS.

1) cVAE: Conditional VAE framework is widely adopted in human scene/object/human interaction generation. We adopt the codes of ACTOR~\cite{actor} as the baseline cVAE model and modify the motion of the actor as a condition. We use the AdamW optimizer with a fixed learning rate of 0.0001 and train the model for 1,000 epochs.
2) AGRoL: We adopt the codes from~\cite{agrol}. The features of the action and reaction sequences are simply concatenated together and fed to the MLP backbone. Same as in AGRoL, we build the MLP network with 12 blocks and the latent dimension is 512. We train the model for 600K steps by the AdamW optimizer with a fixed learning rate of 0.0001.
3) MDM: We adopt the codes from~\cite{mdm} and use the Transformer encoder-only backbone as the baseline. The features of the action and reaction are concatenated together and summed with a standard positional embedding before being fed into the Transformer encoder blocks of 8 layers.  We train the model for 600K steps by the AdamW optimizer with a fixed learning rate of 0.0001.
4) MDM-GRU: We adopt the codes from~\cite{mdm} and use the implemented GRU backbone as the baseline. The features of the action and reaction are added together and fed into the GRU blocks of 8 layers.  We train the model for 600K steps by the AdamW optimizer with a fixed learning rate of 0.0001.
For the text-conditioned setting, we adopt T2M~\cite{guo2022generating}, MDM~\cite{mdm}, MDM-GRU~\cite{mdm}, RAIG~\cite{role_aware} and InterGen~\cite{intergen} as baselines. We also adopt the 6D rotation representation of the SMPL-X parameters as inputs and the frame length is set to 150.

\section{Details of the metric calculations}

For the action-conditioned human reaction generation, we follow the prior works in human motion generation, Action2Motion~\cite{action2motion}, ACTOR~\cite{actor} and MDM~\cite{mdm} to calculate the Frechet Inception Distance(FID)~\cite{fid}, action recognition accuracy, diversity and multi-modality. We borrow the code from the ACTOR~\cite{actor}.
Firstly, we train the action recognition model based on a slightly modified version of ST-GCN~\cite{stgcn}. The ST-GCN model takes the 6D rotation representation of the SMPL-X parameters as input and outputs the action classification results. We train the NTU120-AS and Chi3D-AS datasets for 100 epochs with 64 batch size and 0.0001 learning rate. We generate 20 times of 1000 motion sequences with different random seeds and report the average together with the confidence interval at 95\%. The definition of each metric is as follows:

1) FID: The features are extracted from the generated motions and the real motions. Then the FID is calculated between the feature distribution of the generated motions and the distribution of the real motions;
2) Action recognition accuracy: We use the pre-trained ST-GCN model to classify the generated motions and calculate the accuracy;
3) Diversity: which measures the variance of the generated motions across all action categories. Given the motion feature vectors of generated motions and real motions as \{$v_1, \cdots, v_{S_d}$\} and \{$v_1^\prime, \cdots, v_{S_d}^\prime$\}, the diversity is defined as $Diversity=\frac{1}{S_d}\sum_{i=1}^{S_d}||v_i-v_i^\prime||_2$. $S_d=200$ in our experiments.
4) Multi-modality: which measures how much the generated motions diversify with each action type. Given a set of motions with $C$ action types, for $c$-th action, we randomly sample two subsets with size $S_l$, and then extract the feature vectors as \{$v_{c,1}, \cdots, v_{c, S_l}$\} and \{$v_{c,1}^\prime, \cdots, v_{c,S_l}^\prime$\}, the multimodality is defined as $Multimod.=\frac{1}{C\times S_l}\sum_{c=1}^C\sum_{i=1}^{S_l}||v_{c,i}-v_{c,i}^\prime||_2$. $S_l=20$ in our experiments.

For the text-conditioned human reaction generation, we follow the prior work of~\cite{guo2022generating} and adopt the 1) Frechet Inception Distance (FID)~\cite{fid} to measure the latent distance between real and generated samples, 2) diversity to measure the latent variance, 3) multimodality (MModality) to measure the diversity of the generated results for the same text, 4) R Precision to measure the accuracy of retrieving the ground-truth description from 31 randomly mismatched descriptions, and 5) MultiModal distance (MM Dist) to calculate the latent distance between generated motions and texts. 
We train a motion feature extractor together with a text feature extractor in a contrastive paradigm to align the features of texts and motions.
We run all the evaluations 20 times (except MModality for 5 times) and report the averaged results with the confidence interval at 95\%.

\section{Boarder Impacts}
In AR/VR and gaming applications, a well-trained non-player character (NPC) who can react properly conditioned on your body movements is highly demanded. Our model can be applied to generate plausible human reactions in real time for these applications. 
We believe our work will foster future research in this direction.

\end{document}